\def\year{2020}\relax
\documentclass[letterpaper]{article} % DO NOT CHANGE THIS
\usepackage{aaai20}  % DO NOT CHANGE THIS
\usepackage{times}  % DO NOT CHANGE THIS
\usepackage{helvet} % DO NOT CHANGE THIS
\usepackage{courier}  % DO NOT CHANGE THIS
\usepackage[hyphens]{url}  % DO NOT CHANGE THIS
\usepackage{graphicx} % DO NOT CHANGE THIS
\usepackage{graphicx}  % DO NOT CHANGE THIS
\usepackage{amsfonts}
\usepackage{amsmath}
\usepackage{multirow}
\usepackage{booktabs}
\usepackage{graphicx}
\usepackage{subfigure}
\title{K-BERT: Enabling Language Representation with Knowledge Graph}
\author{\\
Weijie Liu\textsuperscript{\rm 1}, 
Peng Zhou\textsuperscript{\rm 2}, 
Zhe Zhao\textsuperscript{\rm 2}, 
Zhiruo Wang\textsuperscript{\rm 3},
Qi Ju\textsuperscript{\rm 2,*},
Haotang Deng\textsuperscript{\rm 2}
and Ping Wang\textsuperscript{\rm 1,*}\\ 
\textsuperscript{\rm 1}Peking University, Beijing, China \\
\textsuperscript{\rm 2}Tencent Research, Beijing, China\\
\textsuperscript{\rm 3}Beijing Normal University, Beijing, China \\
}
\begin{document}

\maketitle

\begin{abstract}

Pre-trained language representation models, such as BERT, capture a general language representation from large-scale corpora, but lack domain-specific knowledge. When reading a domain text, experts make inferences with relevant knowledge. For machines to achieve this capability, we propose a knowledge-enabled language representation model (K-BERT) with knowledge graphs (KGs), in which triples are injected into the sentences as domain knowledge. However, too much knowledge incorporation may divert the sentence from its correct meaning, which is called knowledge noise (KN) issue. To overcome KN, K-BERT introduces soft-position and visible matrix to limit the impact of knowledge. K-BERT can easily inject domain knowledge into the models by equipped with a KG without pre-training by-self because it is capable of loading model parameters from the pre-trained BERT. Our investigation reveals promising results in twelve NLP tasks. Especially in domain-specific tasks (including finance, law, and medicine), K-BERT significantly outperforms BERT, which demonstrates that K-BERT is an excellent choice for solving the knowledge-driven problems that require experts.

\end{abstract}

\section{Introduction}

% 像BERT这样的无监督预训练语言表示（LR）模型已经在多个NLP任务中取得了非常好的结果。这些模型在大规模语料库上进行预训练，以获得通用语言表示，然后在特定的下游任务中进行微调，例如句子分类，命名实体识别（NER）等。然而，这种模式会导致预训练和下游任务存在领域上的差别，例如，在维基百科上预先培训的Google BERT在处理医疗领域的电子病历（EMR）分析时无法充分发挥其价值。

Unsupervised pre-trained Language Representation (LR) models like BERT \cite{devlin2018bert} have achieved promising results in multiple NLP tasks. These models are pre-trained over large-scale open-domain corpora to obtain general language representations and then fine-tuned in specific downstream tasks for absorbing specific-domain knowledge. However, due to the domain-discrepancy between pre-training and fine-tuning, these models do not perform well on knowledge-driven tasks. For example, the Google BERT pre-trained on Wikipedia can not give full play to its value when dealing with electronic medical record (EMR) analysis task in the medical field.

% 在阅读特定领域的文本时，普通人只能根据其内容进行理解，而专家可以用相关的领域知识进行推理。官方提供的在开放领域语料预训练的模型，如BERT、GPT就相当于一个普通人，尽管能在开放领域任务（如GLUE）上取得很好的效果，但是在特定领域上使用时效果却不尽如人意。为了解决这一问题，业界的常用做法是使用领域语料自己预训练一个模型，替代官方提供的模型。

When reading a text from a specific-domain, ordinary people can only comprehend words according to its context, while experts are able to make inferences with relevant domain knowledge. Publicly-provided models, like BERT, GPT \cite{radford2018improving}, and XLNet \cite{yang2019xlnet}, who were pre-trained over open-domain corpora, act just like an ordinary people. Even though they can refresh the state-of-the-art of GLUE \cite{wang2018glue} benchmark by learning from open-domain corpora, they may fail in some domain-specific tasks, due to little knowledge connection between specific and open domain. One way to solve this problem is to pre-train a model emphasizing domain-specific by ourselves, instead of using the publicly-provided ones. However, pre-training is time-consuming and computationally expensive, making it unacceptable to most users. 

% 虽然使用域语料库进行预训练可以为LR模型提供一些领域知识，但这种知识获取效率非常低且成本高。例如，如果我们希望模型学习“对乙酰氨基酚可以治疗感冒”的知识，我们需要提供大量含有“乙酰氨基酚”和“感冒”共同出现的句子用于预训练。除了从语料库中获取知识外，还有其他方法可以使模型成为领域专家吗？

Moreover, although injecting domain-specific knowledge during pre-training is possible for LR models, this process of knowledge acquisition can be inefficient and expensive. For example, if we want the model to acquire the knowledge of ``Paracetamol can treat cold", a large number of co-occurrences of "Paracetamol" and "cold" are required in the pre-training corpus. Instead of this strategy, what else can we do to make the model a domain expert? The knowledge graph (KG), which was called ontology in early research, serves as a good solution. As knowledge refined into a structured form, KGs over many fields have been constructed, e.g., SNOMED-CT \cite{bodenreider2008biomedical} in medical field, HowNet \cite{dong2006hownet} in Chinese conception. If KG can be integrated into the LR model, it will equip the model with domain knowledge, enhancing the model's performance over domain-specific tasks, while reducing the cost of pre-training on a large scale. Besides, the resulting models posses greater interpretability, for the injected knowledge is manually editable.

% 然而，将知识图谱融合到语言模型中存在许多挑战：（1）异构嵌入空间（HES）：通常，文本中的单词嵌入向量和KG中的实体以不同的方式获得，从而使向量空间不一致。(2) 知识噪声：为句子添加太多知识可能会改变其原始含义。为了克服这些挑战，我们提出了一种知识使能双向编码器表示（K-BERT）。K-BERT可以直接加载Google Inc.提供的预训练模型，并加载KG以获得领域知识而无需预先培训。K-BERT的这一特性对于计算资源有限的用户来说非常方便。12个中文NLP任务的实验结果表明，K-BERT与KG在领域任务方面表现出色。本文的主要贡献可归纳如下：
% 1. 我们提出了一个知识使能的LR模型，即K-BERT，它可以在没有HES和KN问题的情况下为文本添加一般或领域知识;
% 2. 在KG的帮助下，K-BERT在各种开放或特定领域的NLP任务上明显优于BERT;
% 3. 我们发布了K-BERT的代码和我们自行开发的知识图谱，可以在https://github.com/xxxx/xxx/xxx中找到.

However, there are two challenges lies in the road of this knowledge integration: \textbf{(1)} \textbf{Heterogeneous Embedding Space} (HES): In general, the embedding vectors of words in text and entities in KG are obtained in separate ways, making their vector-space inconsistent; \textbf{(2)} \textbf{Knowledge Noise} (KN): Too much knowledge incorporation may divert the sentence from its correct meaning. To overcome these challenges, this paper propose a \textbf{K}nowledge-enabled \textbf{B}idirectional \textbf{E}ncoder \textbf{R}epresentation from \textbf{T}ransformers (K-BERT). K-BERT is capable of loading any pre-trained BERT models due to they are identical in parameters. In addition, K-BERT can easily inject domain knowledge into the models by equipped with a KG without pre-training. This characteristic of K-BERT is very convenient for users with limited computing resources. Experimental results on twelve Chinese NLP tasks demonstrate that the K-BERT gains superior performances on domain-specific tasks. The main contributions of this paper can be summarized as follows:

\begin{itemize}
\item This paper proposes a knowledge-enabled LR model, namely K-BERT, which is compatible with BERT and can incorporate domain knowledge without HES and KN issues;
\item With the delicate injection of KG, K-BERT significantly outperforms BERT not only on domain-specific tasks, but also plenty of those in the open-domain;
\item The codes of K-BERT and our self-developed knowledge graphs are publicly available at \url{https://github.com/autoliuweijie/K-BERT}.
\end{itemize}

\section{Related Work}
\label{sec:related_work}

Since Google Inc. launched BERT in 2018, many endeavors have been made for further optimization, basically focusing on the pre-training process and the encoder.

In optimizing pre-training process, Baidu-ERNIE \cite{sun2019ernie} and BERT-WWM \cite{cui2019pre} adopt whole-word masking rather than single character masking for pre-training BERT in Chinese corpora. SpanBERT \cite{joshi2019spanbert} extended BERT by masking contiguous random spans and proposed a span boundary objective. RoBERTa \cite{liu2019roberta} optimized the pre-training of BERT in three ways, i.e., deleting the target of the next sentence prediction, dynamically changing the masking strategy and using more and longer sentences for training. In optimizing the encoder of BERT, XLNet \cite{yang2019xlnet} replaced the Transformer in BERT with Transformer-XL \cite{dai2019transformer} to improve its ability to process long sentences. THU-ERNIE \cite{zhang2019ernie} modified the encoder of BERT to an aggregator for the mutual integration of word and entities.

% 由于预训练LR模型是一个新兴的方向，因此很少有关于它与KG融合的研究。THU-ERNIE是这方面的先驱，但它只包含实体信息而忽略了KG中的关系。COMET使用KG中的三元组来训练GPT，使其学会常识。在预训练语言模型兴起之前，有一些研究将知识图谱与词向量表示相结合。

While the pre-trained LR model is an emerging direction, there is little work on its fusion with KG. THU-ERNIE \cite{zhang2019ernie} is a pioneer in this direction by fusing entity information, but the relations between entities are ignored by it. COMET \cite{bosselut2019comet} employed the triples in KG as corpus to train GPT \cite{radford2018improving} for common sense learning, which is very inefficient. 

Before the emergence of pre-trained LR models, there were several studies that combined KG with word vectors. \citeauthor{wang2014knowledge} (\citeyear{wang2014knowledge}) proposed a novel method of jointly embedding entities and words into the same continuous vector space basing on the idea of word2vec \cite{mikolov2013efficient}. \citeauthor{toutanova2015representing} (\citeyear{toutanova2015representing}) proposed a model that captures the compositional structure of textual relations, and optimize entity, knowledge base, and textual relation representations in a joint manner. \citeauthor{han2016joint} (\citeyear{han2016joint}) applied a convolutional neural network and a KG completion task to learn the representation of text and knowledge jointly. \citeauthor{cao2018joint} (\citeyear{cao2018joint}) carried out cross-lingual representation learning for words and entities via attentive distant supervision.

The major weakness of these methods is that they are still based on the idea of ``word2vec + transE" \cite{bordes2013translating}, rather than the pre-trained LR model. Although they use the method of joint representation to make the vector space of entities and words closer, there are still HES problems. What's more, for KGs with millions of entities, this idea makes the entity table very large, making it unusable  because it exceeds the GPU's memory size.

\begin{figure}[]
\centering
\includegraphics[width=0.9\columnwidth]{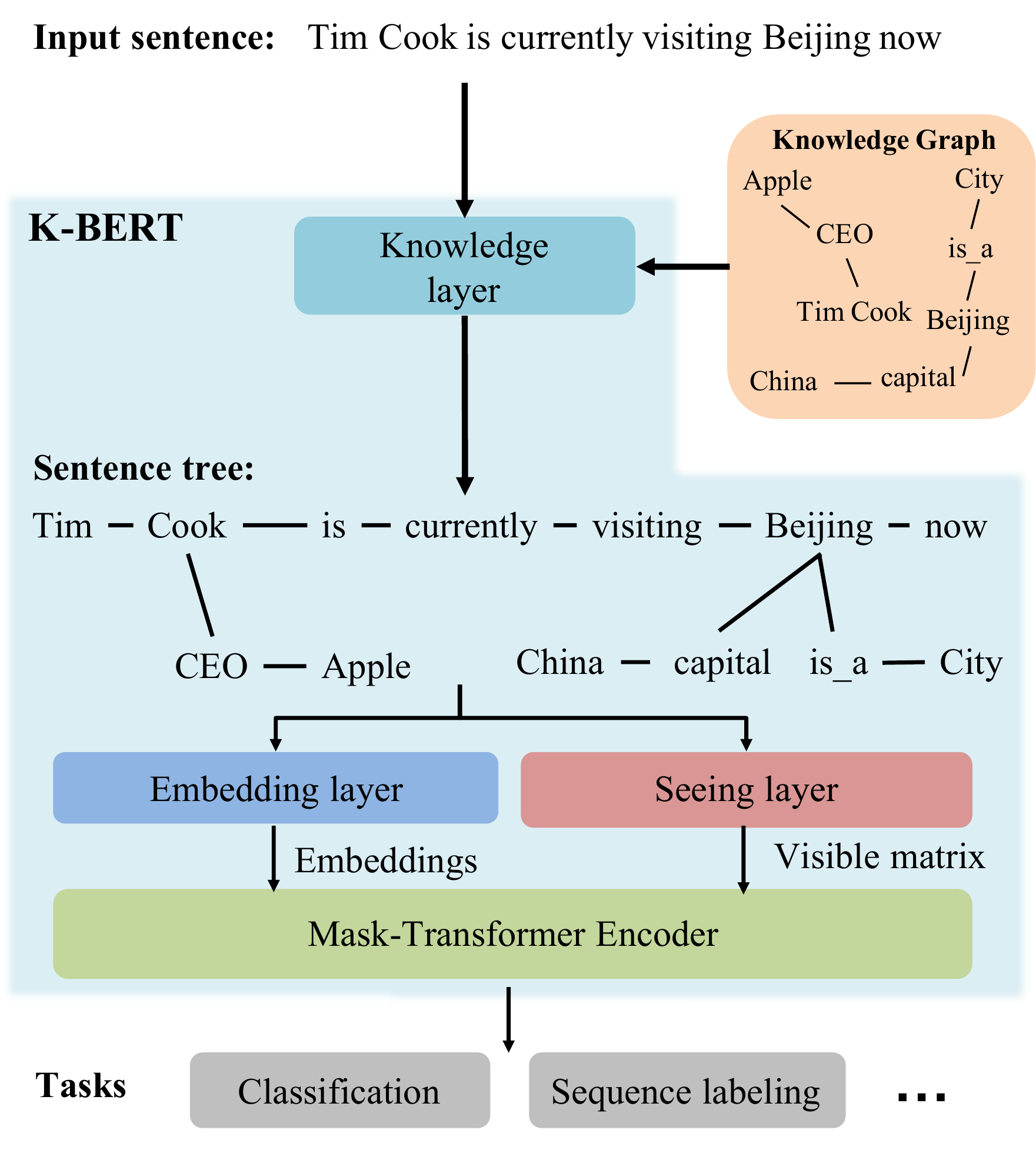}
\caption{The model structure of K-BERT: Compared to other RL models, the K-BERT is equipped with an editable KG, which can be adapted to its application domain. For example, for electronic medical record analysis, we can use a medical KG to grant the K-BERT with medical knowledge.}
\label{model_structure}
\end{figure}

\begin{figure*}[t]
\centering
\includegraphics[width=7in]{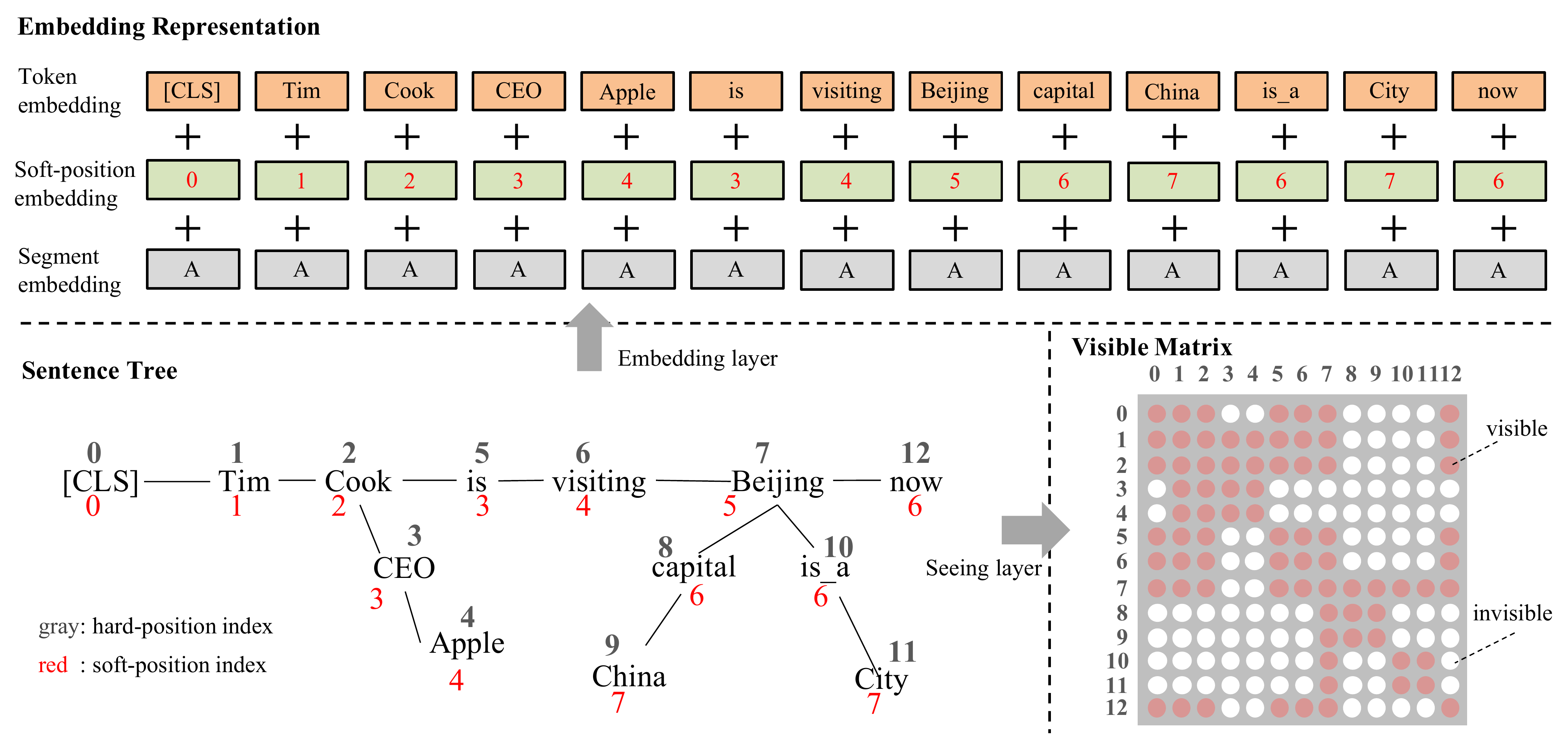}
\caption{The process of converting a sentence tree into an embedding representation and a visible matrix. In the sentence tree, the red number is the soft-position index, and the gray is the hard-position index. \textbf{(1)} For token embedding, the tokens in the sentence tree are flattened into a sequence of token embedding by their hard-position index; \textbf{(2)} The soft-position index is used as position embedding along with the token embedding; \textbf{(3)} In segment embedding, all the tokens in the fist sentence are tagged as ``A"; \textbf{(4)} In the visible matrix, red means visible, and white means invisible. For example, the cell at row 4, column 9 is white means that the "Apple(4)" cannot see "China(9)".}
\label{embedding_seeing}
\end{figure*}

\section{Methodology}
\label{sec:methodology}

In this section, we detail the implementation of K-BERT and its overall framework is presented in Figure \ref{model_structure}. 

\subsection{Notation}

We denote a sentence $s = \{w_0, w_1, w_2, ..., w_n\}$ as a sequence of tokens, where $n$ is the length of this sentence. In this paper, English tokens are taken at the word-level, while Chinese tokens are at character-level. Each token $w_i$ is included in the vocabulary $\mathbb{V}$, $w_i\in\mathbb{V}$. KG, denoted as $\mathbb{K}$, is a collection of triples $\varepsilon = (w_i, r_j, w_k)$, where $w_i$ and $w_k$ are the name of entities, and $r_j\in\mathbb{V}$ is the relation between them. All the triples are in KG, i.e., $\varepsilon \in \mathbb{K}$.

\subsection{Model architecture}

% 如图1所示，K-BERT的整个模型体系结构由四个模块组成，即知识层，嵌入层，视域层和掩模变换器。对于输入句子，知识层首先基于KG将相关三元素注入其中，将原始句子转换为知识丰富的句子树。然后将句子树同时输入到嵌入层和视觉层中，以便转换为令牌级嵌入表示和可见矩阵。 可见矩阵用于控制每个令牌的可见区域，避免由于注入过多的知识而改变原始句子的含义。

As shown in Figure \ref{model_structure}, the model architecture of K-BERT consists of four modules, i.e., knowledge layer, embedding layer, seeing layer and mask-transformer. For an input sentence, the knowledge layer first injects relevant triples into it from a KG, transforming the original sentence into a knowledge-rich sentence tree. The sentence tree is then simultaneously fed into the embedding layer and the seeing layer and then converted to a token-level embedding representation and a visible matrix. The visible matrix is used to control the visible area of each token, preventing changing the meaning of the original sentence due to too much knowledge injected. 

\subsection{Knowledge layer}

The knowledge layer (KL) is used for sentence knowledge injection and sentence tree conversion. Specifically, given an input sentence $s=\{w_0, w_1, w_2, ..., w_n\}$ and a KG $\mathbb{K}$, KL outputs a sentence tree $t=\{w_0, w_1, ..., w_i \{(r_{i0}, w_{i0}), ..., (r_{ik}, w_{ik})\}, ..., w_n\}$. This process can be divided into two steps: knowledge query (K-Query) and knowledge injection (K-Inject).

In K-Query, all the entity names involved in the sentence $s$ are selected out to query their corresponding triples from $\mathbb{K}$. K-Query can be formulated as (\ref{k_query}),
\begin{equation}
\label{k_query}
E=K\_Query(s, \mathbb{K}),
\end{equation}
where $E=\{(w_{i}, r_{i0}, w_{i0}), ..., (w_{i}, r_{ik}, w_{ik})\}$ is a collection of the corresponding triples. 

\begin{figure}[htb]
\centering
\includegraphics[width=0.8\columnwidth]{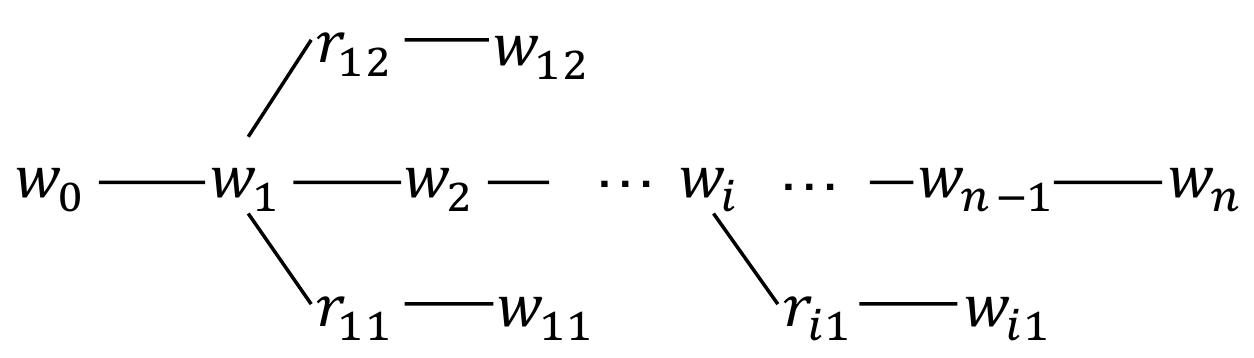} 
\caption{Structure of the sentence tree.}
\label{sentence_tree}
\end{figure}

Next, K-Inject injects the queried $E$ into the sentence $s$ by stitching the triples in $E$ to their corresponding position, and generates a sentence tree $t$. The structure of $t$ is illustrated in Figure \ref{sentence_tree}.

In this paper, a sentence tree can have multiple branches, but its depth is fixed to $1$, which means that the entity names in triples will not derive branches iteratively. K-Inject can be formulated as (\ref{k_inject}),
\begin{equation}
\label{k_inject}
t=K\_Inject(s, E).
\end{equation}

\subsection{Embedding layer}

The function of the Embedding Layer (EL) is to convert the sentence tree into an embedding representation that can be fed into the Mask-Transformer. Similar to BERT, the embedding representation of K-BERT is the sum of three parts: token embedding, position embedding, and segment embedding, but differs in that the input of the K-BERT's embedding layer is a sentence tree, rather than a token sequence. Therefore, how to transform a sentence tree into a sequence while retaining its structural information is the key to K-BERT.

\subsubsection{Token embedding} In this work, the token embedding is consistent with BERT, and the vocabulary provided by Google BERT is adopted in this paper. Each token in the sentence tree is converted to an embedding vector of dimension $H$ via a trainable lookup table. In addition, K-BERT also uses $[CLS]$ as a classification tag and uses $[MASK]$ to mask tokens. The difference between the token embeddings of K-BERT and BERT is that the tokens in the sentence tree require re-arrangement before the embedding operation. In our re-arrange strategy, tokens in the branch are inserted after the corresponding node, while subsequent tokens are moved backwards. As the example illustrated in Figure \ref{embedding_seeing}, the sentence tree is rearranged as ``Tim Cook CEO Apple is visiting Beijing capital China is\_a City now". Although this procedure is simple, but it makes the sentence unreadable and lost correct structural information. Fortunately, it can be solved by the soft-position and visible matrix.

\subsubsection{Soft-position embedding} For BERT, if there is no position embedding, it will be equivalent to a bag-of-word model, resulting in a lack of structural information (i.e., the order of tokens). All the structural information of the BERT's input sentence is contained in the position embedding, which allows us to add the missing structural information back to the unreadable rearranged sentence. Taking the sentence tree in Figure \ref{embedding_seeing} as an example, after rearranging, $[CEO]$ and $[Apple]$ are inserted between $[Cook]$ and $[is]$, but the subject of $[is]$ should be $[Cook]$ instead of $[Apple]$. To solve this problem, we only need to set the position number of $[is]$ to $3$ instead of $5$. So when calculating the self-attention score in the transformer encoder, $[is]$ is at the next position of $[Cook]$ by the equivalent. However, another problem arises: the position numbers of $[is]$ and $[CEO]$ are both $3$, which makes them close in position when calculating self-attention, but in fact, there is no connection between them. The solution to this problem is Mask-Self-Attention, which will be covered in the next subsection.

\subsubsection{Segment embedding} Like BERT, K-BERT also uses segmentation embedding to identify different sentences when multiple sentences are included. For example, when two sentences $\{w_{00}, w_{01}, ..., w_{0n}\}$ and $\{w_{10}, w_{11}, ..., w_{1m}\}$ are inputed, they are combined into one sentence  $\{[CLS], w_{00}, w_{01}, ..., w_{0n}, [SEP], w_{10}, w_{11}, ..., w_{1m}\}$  with $[SEP]$. For the combined sentence, it is marked with a sequence of segment tags, $\{A, A, A, A, ..., A, B, B, ..., B\}$.

\subsection{Seeing layer}
\label{seeing_layer}

Seeing layer is the biggest difference between K-BERT and BERT, and also what make this method so effective. The input to K-BERT is a sentence tree, where the branch is the knowledge gained from KG. However, the risk raised with knowledge is that it can lead to changes in the meaning of the original sentence, i.e., KN issue. For example, in the sentence tree in Figure \ref{embedding_seeing}, $[China]$ only modifies $[Beijing]$ and has nothing to do with $[Apple]$. Therefore, the representation of $[Apple]$ should not be affected by $[China]$. On the other hand, the $[CLS]$ tag used for classification should not bypass the $[Cook]$ to get the information of $[Apple]$, as this would bring the risk of semantic changes. To prevent this from happening, K-BERT's use a visible matrix $M$ to limit the visible area of each token so that $[Apple]$ and $[China]$, $[CLS]$ and $[Apple]$ are not visible to each other. The visible matrix $M$ is defined as (\ref{eq_visible_matrix}),
\begin{equation}
\label{eq_visible_matrix}
M_{ij}=\left\{\begin{matrix}
0 &  w_i \ominus  w_j \\ 
 -\infty  & w_i \oslash  w_j
\end{matrix}\right.
\end{equation}
where, $w_i \ominus  w_j$ indicates that $w_i$ and $w_j$ are in the same branch, while $w_i \oslash  w_j$ are not. $i$ and $j$ are the hard-position index.

\subsection{Mask-Transformer}

\begin{figure}[t]
\centering
\includegraphics[width=0.9\columnwidth]{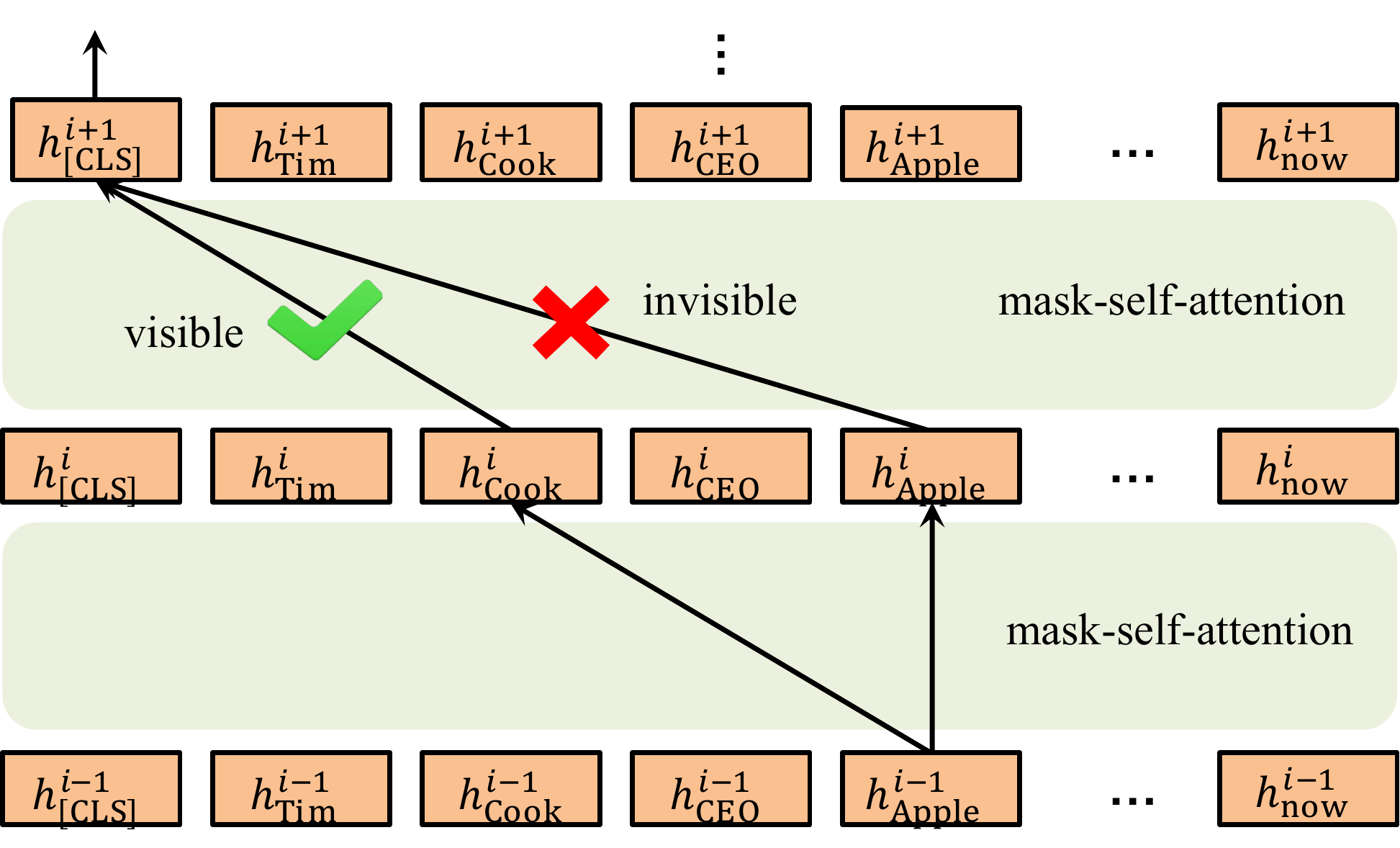}
\caption{Illustration of the Mask-Transformer, which is a stack of multiple mask-self-attention blocks.}
\label{figure_mask_self_attention}
\end{figure}

To some degree, the visible matrix $M$ contains the structural information of the sentence tree. The Transformer \cite{vaswani2017attention} encoder in BERT cannot receive $M$ as an input, so we need to modify it to Mask-Transformer, which can limit the self-attention region according to $M$. Mask-Transformer is a stack of multiple mask-self-attention blocks. As BERT, we denote the number of layers (i.e., mask-self-attention blocks) as $L$, the hidden size as $H$, and the number of mask-self-attention heads as $A$.

\subsubsection{Mask-Self-Attention:}
\label{mask_self_attention}

To prevent false semantic changes by taking advantage of the sentence structure information in $M$, we propose a mask-self-attention, which is an extension of self-attention. Formally, the mask-self-attention is (\ref{eq_msak_self_attention}).

\begin{equation}
\label{eq_msak_self_attention}
Q^{i+1},\, K^{i+1},\, V^{i+1} = h^{i}W_q,\, h^{i}W_k,\,  h^{i}W_v,
\end{equation}
\begin{equation}
S^{i+1} = softmax(\frac{Q^{i+1}{K^{i+1}}^{\top} + M}{\sqrt{d_k}}),
\end{equation}
\begin{equation}
\label{eq_msak_self_attention_2}
h^{i+1} = S^{i+1}V^{i+1},
\end{equation}
where $W_q$, $W_k$ and $W_v$ are trainable model parameters. $h^{i}$ is the hidden state of the $i$-th mask-self-attention blocks. $d_k$ is the scaling factor\footnote{In \cite{vaswani2017attention}, the author scale the dot products by $\frac{1}{\sqrt{d_k}}$ to counteract the effect of the dot products grow large in magnitude.}. $M$ is the visible matrix calculated by the seeing layer. Intuitively, if $w_k$ is invisible to $w_j$, the $M_{jk}$ will mask the attention score $S^{i+1}_{jk}$ to $0$, which means $w_k$ make no contribution to the hidden state of $w_j$.

As the example illustrated in Figure \ref{figure_mask_self_attention}, $h^{i}_{[Apple]}$ has no effect on $h^{i+1}_{[CLS]}$, because $[Apple]$ is invisible to $[CLS]$. However, $h^{i+1}_{[CLS]}$ can obtain the information of $h^{i-1}_{[Apple]}$ indirectly through $h^{i+1}_{[Cook]}$, because $[Apple]$ is visible to $[Cook]$ and $[Cook]$ is visible to $[CLS]$. The advantage of this process is that $[Apple]$ enriches the representation of $[Cook]$, but does not directly affect the meaning of the original sentence.

\section{Experiments}
\label{sec:experiments}

In this section, we present the details of the K-BERT fine-tuning results on twelve Chinese NLP tasks, among which eight are open-domain, and four are specific-domain.

\begin{table*}[]
\small
\caption{Results of various models on sentence classification tasks on open-domain tasks ($Acc.$ \%)}
\label{table_open_domain}
\center
\begin{tabular}{lcccccccccccc}
\toprule
\multirow{2}{*}{\textbf{Models\textbackslash{}Datasets}} & \multicolumn{2}{c}{\textbf{Book\_review}} & \multicolumn{2}{c}{\textbf{Chnsenticorp}} & \multicolumn{2}{c}{\textbf{Shopping}} & \multicolumn{2}{c}{\textbf{Weibo}} & \multicolumn{2}{c}{\textbf{XNLI}} & \multicolumn{2}{c}{\textbf{LCQMC}} \\  
                                                         & $Dev$                 & $Test$                 & $Dev$                 & $Test$                 & $Dev$               & $Test$               & $Dev$              & $Test$             & $Dev$             & $Test$             & $Dev$              & $Test$             \\ \midrule
\multicolumn{13}{c}{Pre-trainied on WikiZh by Google.}                                                                                                                                                                                                                                                                \\ \midrule
Google BERT                                              & 88.3                 & \textbf{87.5}         & 93.3                & 94.3                 & 96.7             & 96.3                & 98.2             & 98.3             & 76.0                & 75.4                 & 88.4                 & 86.2              \\ \midrule
K-BERT (HowNet)                                          & 88.6                 & 87.2                   & \textbf{94.6}          & \textbf{95.6}        & \textbf{97.1}   & \textbf{97.0}     & 98.3                 &  98.3                & \textbf{76.8}        & \textbf{76.1}      & \textbf{88.9}     & 86.9             \\ 
K-BERT (CN-DBpedia)                                      & \textbf{88.6}                 &  87.3                    & 93.9             & 95.3                     & 96.6              &  96.5             &  98.3                & 98.3                  & 76.5          & 76.0           & 88.6              & \textbf{87.0}          \\ \midrule
\multicolumn{13}{c}{Pre-trained on WikiZh and WebtextZh by us.}                                                                                                                                                                                                                                                      \\ \midrule
Our BERT                                                 & \textbf{88.6}       & 87.9                 & 94.8                    & 95.7                 & 96.9             & \textbf{97.1}      & 98.2             & 98.2             & 77.0                & 76.3                 & 89.0             & 86.7             \\ \midrule
K-BERT (HowNet)                                          & 88.5                & 87.4                 & \textbf{95.4}                & 95.6                 & 96.9             & 96.9               & 98.3             & \textbf{98.4} & \textbf{77.2}                & \textbf{77.0}             & \textbf{89.2}          & \textbf{87.1}        \\
K-BERT (CN-DBpedia)                                      & 88.8                & 87.9                 & 95.0                    & \textbf{95.8}        & \textbf{97.1}    & 97.0               & 98.3             & 98.3             & 76.2            & 75.9             & 89.0             & 86.9             \\ \bottomrule
\end{tabular}
\end{table*}

\begin{table}[]
\small
\center
\caption{Results of various models on NLPCC-DBQA ($MRR$ \%) and MSRA-NER ($F1$ \%).}
\label{table_dbqa_msra}
\begin{tabular}{lcccc}
\toprule
\multirow{2}{*}{\textbf{Models\textbackslash{}Datasets}} & \multicolumn{2}{l}{\textbf{NLPCC-DBQA}} & \multicolumn{2}{l}{\textbf{MSRA-NER}} \\ 
                                                         & $Dev$       & $Test$       & $Dev$      & $Test$      \\ \midrule
\multicolumn{5}{c}{Pre-trained on WikiZh by Google.}                                                                                                 \\ \midrule
Google BERT                                              & 93.4              & 93.3               & 94.5             & 93.6             \\ \midrule
K-BERT (HowNet)                                          & 93.2              & 93.1               & 95.8             & 94.5                    \\
K-BERT (CN-DBpedia)                                      & \textbf{94.5}     & \textbf{94.3}      & \textbf{96.6}        & \textbf{95.7}           \\ \midrule
\multicolumn{5}{c}{Pre-trained on WikiZh and WebtextZh by us.}                                                                                       \\ \midrule
Our BERT                                                 & 93.3               & 93.6                & 95.7             & 94.6               \\ \midrule
K-BERT (HowNet)                                          & 93.2               & 93.1                & 96.3             & 95.6               \\
K-BERT (CN-DBpedia)                                      & \textbf{93.6}      & \textbf{94.2}       & \textbf{96.4}              & \textbf{95.6}      \\ \bottomrule
\end{tabular}
\end{table}

\subsection{Pre-training corpora}

In this paper, we adopt two Chinese corpora for pre-training, i.e., WikiZh\footnote{\url{https://dumps.wikimedia.org/zhwiki/latest/}} and WebtextZh\footnote{\url{https://github.com/brightmart/nlp_chinese_corpus}}.

\begin{itemize}
\item \textbf{WikiZh} WikiZh refers to the Chinese Wikipedia corpus, which is used to train Chinese BERT in \cite{devlin2018bert}.  WikiZh contains a total of 1 million well-formed Chinese entries with 120 million sentences and size of 1.2G.

\item \textbf{WebtextZh} WebtextZh is a large-scale, high-quality Chinese question and answer (Q\&A) corpus with 4.1 million entries and a size of 3.7G. Each entry in WebtextZh belongs to a topic, with a total of 28,000 topics.
\end{itemize}

\subsection{Knowledge graph}

We employ three Chinese KGs, CN-DBpedia\footnote{\url{http://kw.fudan.edu.cn/cndbpedia/intro/}}, HowNet\footnote{\url{http://www.keenage.com/}} and MedicalKG.

\begin{itemize}
\item \textbf{CN-DBpedia} CN-DBpedia \cite{xu2017cn-dbpedia} is a large-scale open-domain encyclopedic KG developed by the Knowledge Work Laboratory of Fudan University, covering tens of millions of entities and hundreds of millions of relations. In this paper, we refine the official CN-DBpedia by eliminating those triples whose entity names are less than 2 in length or contain special characters. The refined CN-DBpedia contains a total of 5.17 million triples.

\item \textbf{HowNet} HowNet is a large-scale language knowledge base for Chinese vocabulary and concepts \cite{dong2006hownet}, in which each Chinese word is annotated with semantic units called sememes. If we take \{word, contain, sememes\} as a triple, HowNet is a language KG. Similarly, we refine the official HowNet by eliminating those triples whose entity names are less than 2 in length or contain special characters. The refined HowNet contains a total of 52,576 triples.

\item \textbf{MedicalKG} MedicalKG is our self-developed Chinese medical concept KG, which contains four types of hypernym (symptoms, diseases, parts, and treatments). MedicalKG contains a total of 13,864 triples and is open source as part of K-BERT.
\end{itemize}

\subsection{Baselines}

In this paper, we compare K-BERT to two baselines:

\begin{itemize}
\item \textbf{Google BERT}\footnote{\url{https://github.com/google-research/bert}} The model was pre-trained on WikiZh and released by Google \cite{devlin2018bert}. 
\item \textbf{Our BERT} Our reimplementation of BERT with pre-training on WikiZh and WebtextZh.
\end{itemize}

\subsection{Parameter settings and training details}

To reflect the role of KG in the RL model, we configure our K-BERT and BERT to the same parameter settings according to the basic version of Google BERT \cite{devlin2018bert}. We denote the number of (mask-)self-attention layers and heads as $L$ and $A$ respectively, and the hidden dimension of embedding vectors as $H$. In detail, we have the following model configuration: $L=12$, $A=12$ and $H=768$. The total amounts of trainable parameters of both BERT and K-BERT are the same (110M), which means that they are compatible with each other in model parameters.

For K-BERT pre-training, all settings are consistent with \cite{devlin2018bert}. One thing to emphasize is that we don't add any KG to K-BERT during the pre-training phase. Because KG binds two related entity names together, thus making the pre-trained word vectors of the two are very close or even equal and resulting in a semantic loss. Therefore, in the pre-training phase, K-BERT and BERT are equivalent, and the latter's parameters can be assigned to the former. KG will be enabled during the fine-tuning and inferring phases.

\subsection{Open-domain tasks}

In this paper, we first compare the performance of K-BERT with the BERT on eight Chinese open-domain NLP tasks. Among these eight tasks, Book\_review\footnote{\url{https://embedding.github.io/evaluation/}}, Chnsenticorp\footnote{\url{https://github.com/pengming617/bert_classification}}, Shopping\footnote{\url{https://share.weiyun.com/5xxYiig}}, and Weibo\footnote{\url{https://share.weiyun.com/5lEsv0w}} are single-sentence classification tasks:
\begin{itemize}
\item \textbf{Book\_review} This dataset contains 20,000 positive and 20,000 negative reviews collected from Douban\footnote{\url{https://book.douban.com/}};
\item \textbf{Chnsenticorp} Chnsenticorp is a hotel review dataset with a total of 12,000 reviews, including 6,000 positive reviews and 6,000 negative reviews;
\item \textbf{Shopping} Shopping is a online shopping review dataset that contains 40,000 reviews, including 21,111 positive reviews and 18,889 negative reviews;
\item \textbf{Weibo} Weibo is a dataset with emotional annotations from Sina Weibo, including 60,000  positive samples and 60,000 negative samples.
\end{itemize}
XNLI \cite{conneau2018xnli}, LCQMC \cite{liu2018lcqmc} are two-sentence classification tasks, NLPCC-DBQA\footnote{\url{http://tcci.ccf.org.cn/conference/2016/dldoc/evagline2.pdf}} is a Q\&A matching task, and MSRA-NER \cite{Levow2006The} is a Named Entity Recognition (NER) task:
\begin{itemize}
\item \textbf{XNLI} XNLI is a cross-language language understanding dataset in which each entry contains two sentences and the task is to determine their relation (``Entailment", ``Contradict" or ``Neutral" );
\item \textbf{LCQMC} LCQMC is a large-scale Chinese question matching corpus. The goal of this task is to determine if the two questions have a similar intent; 
\item \textbf{NLPCC-DBQA} NLPCC-DBQA is a task to predict answers to each question from the given document; 
\item \textbf{MSRA-NER} MSRA-NER is a NER dataset published by Microsoft. This task is to recognize the entity names in the text, including person names, place names, organization names, etc.
\end{itemize}

\begin{table*}[]
\small
\center
\caption{Results of various models on specific-domain tasks (\%).}
\label{table_domain}
\begin{tabular}{l|ccc|ccc|ccc|ccc}
\toprule
\multicolumn{1}{c}{\multirow{2}{*}{\textbf{Models\textbackslash{}Datasets}}} & \multicolumn{3}{c}{\textbf{Finance\_Q\&A}} & \multicolumn{3}{c}{\textbf{Law\_Q\&A}} & \multicolumn{3}{c}{\textbf{Finance\_NER}} & \multicolumn{3}{c}{\textbf{Medicine\_NER}} \\  
\multicolumn{1}{c|}{}                                                         & $P.$            & $R.$           & $F1$           & $P.$          & $R.$          & $F1$          & $P.$           & $R.$           & $F1$           & $P.$           & $R.$           & $F1$           \\ \midrule
\multicolumn{13}{c}{Pre-trained on WikiZh by Google.}                                                                                                                                                                                                                     \\  \midrule
Google BERT                                                                    & 81.9         & 86.0        & 83.9        & 83.1       & 90.1       & 86.4       & 84.8             & 87.4             & 86.1             & 91.9             & 93.1            & 92.5              \\ \midrule
K-BERT (HowNet)                                                                & 83.3         & 84.4        & 83.9        & 83.7       & 91.2       & 87.3       & 86.3        & 89.0        & \textbf{87.6}   & 93.2        & 93.3         & 93.3         \\
K-BERT (CN-DBpedia)                                                            & 81.5          & 88.6        & \textbf{84.9} & 82.1       & 93.8       & \textbf{87.5}  & 86.1        & 88.7        & 87.4        &  93.9       & 93.8         & 93.8          \\
K-BERT (MedicalKG)                                                             & -             & -            & -            & -           & -           & -           & -            & -            & -            & 94.0             & 94.4              & \textbf{94.2}     \\  \midrule
\multicolumn{13}{c}{Pre-trained on WikiZh and WebtextZh by us.}                                                                                                                                                                                                           \\  \midrule
Our BERT                                                                       & 82.1          & 86.5         & 84.2         & 83.2        & 91.7        & 87.2        & 84.9         & 87.4         & 86.1         & 91.8         & 93.5            & 92.7          \\ \midrule
K-BERT (HowNet)                                                                & 82.8          & 85.8         & 84.3         & 83.0        & 92.4        & 87.5        & 86.3         & 88.5         & 87.3         & 93.5         & 93.8             & 93.7          \\
K-BERT (CN-DBpedia)                                                            & 81.9         & 87.1        & \textbf{84.4} & 83.1        & 92.6        &\textbf{87.6} & 86.3         & 88.6         & \textbf{87.4}  & 93.9         & 94.3           & 94.1          \\
K-BERT (MedicalKG)                                                             & -             & -            & -            & -           & -           & -           & -            & -            & -            &  94.1        & 94.3           & \textbf{94.2} \\ \bottomrule
\end{tabular}
\end{table*}

% 以上数据集被分为三个部分：train，dev和test。 我们使用train部分来微调模型，然后在dev和test中测试模型的性能。实验结果显示在表1和表2中，其结果可分为三类: (1) 在句子情感分类任务中，KG对情感分析任务的任务没有显着影响，因为句子的情感可以基于没有任何知识的情感词来判断；（2）对于语义相似性任务，语言类知识图谱（HowNet）的效果优于百科类知识图谱；（3）对于问答和NER任务，百科类知识图谱比语言类知识图谱更合适。 由此，根据任务的类型选择合适的KG是十分重要的。

Each of the above datasets is divided into three parts: $train$, $dev$, and $test$. We use the $train$ part to fine-tune the model and then evaluate its performance on the $dev$ and $test$ parts. The experimental results are shown in Table \ref{table_open_domain} and \ref{table_dbqa_msra}, from which the results can be divided into three categories: \textbf{(1)} \textbf{The KG has no significant effect on the tasks of sentiment analysis} (i.e., Book\_review, Chnsenticorp, Shopping and Weibo) because the sentiment of a sentence can be judged based on emotional words without any knowledge; \textbf{(2)} \textbf{The language KG (HowNet) performs better than the encyclopedic KG in terms of semantic similarity tasks} (i.e., XNLI and LCQMC); \textbf{(3)} For Q\&A and NER tasks (i.e., NLPCC-DBQA and MSRA-NER), \textbf{the encyclopedic KG (CN-DBpedia) is more suitable than the language KG.} Therefore, it is important to choose the right KG based on the type of task.

% 另外，我们发现，对于大多数任务使用额外的语料（WebtextZh）也能带来一定的性能提升，但是其提升效果没有使用知识图谱明显。正如表2中显示的MSRA-NER，CN-DBpedia将F1从93.6提高到95.7，而WebtextZh仅将其提高到94.6。

In addition, it can be observed that the use of an additional corpus (WebtextZh) can also bring performance boost, but not as significant as KG. As MSRA-NER shown in Table \ref{table_dbqa_msra}, the CN-DBpedia improves $F1$ from $93.6\%$ to $95.7\%$, while the WebtextZh only increases it to $94.6\%$.

\subsection{Specific-domain tasks}

In fact, the task where K-BERT really shines is in specific-domain. Because KG is good at giving LR model with domain knowledge. 
\subsubsection{Domain Q\&A} We crawl about 770,000 and 36,000 Q\&A samples from Baidu Zhidao\footnote{\url{https://zhidao.baidu.com}} in financial and legal domains, including questions, netizen answers, and best answers. Based on this, we built two datasets, i.e., \textbf{Finance\_Q\&A} and \textbf{Law\_Q\&A}. The task is to choose the best answer for the question from the netizen's answers.

\subsubsection{Domain NER} \textbf{Finance\_NER}\footnote{\url{https://embedding.github.io/evaluation/#extrinsic}} is a dataset including 3000 financial news articles manually labeled with over 65,000 name entities (people, location and organization). \textbf{Medicine\_NER} is the Clinical Named Entity Recognition (CNER) task released in CCKS 2017\footnote{\url{https://biendata.com/competition/CCKS2017_2/}}. The goal is to extract medical-related entity names from electronic medical records.

% 同样，特定域数据集分为三个部分：train，dev和test，分别用于微调，选择和测试模型。各种模型的测试结果在表3中说明，其中P.，R.和F1分别指Precision，Recall和F1-score。与BERT相比，K-BERT在领域任务方面有明显的性能提升。 至于F1，带有CN-DBpedia的K-BERT可以将所有任务的性能提高1~2％。这独特的收益受益于KG的领域知识。此外，从表3中的Medicine_NER可以观察到使用医学领域KG MedicalKG的性能改善非常明显。 从这些结果中，我们可以得出这样的结论：KG，特别是域KG，对域任务非常有帮助。

Similarly, the specific-domain datasets are split into three parts: $train$, $dev$, and $test$, which are used to fine-tune, select and test model, respectively. The test results of various models are illustrated in Table \ref{table_domain}, where $P.$, $R.$ and $F1$ refer to Precision, Recall and F1-score, respectively. Compared with BERT, K-BERT has a significant performance improvement in terms of domain tasks. As for $F1$, K-BERT with CN-DBpedia can improve the performance of all tasks by 1$\sim$2\%. The unique gain benefits from the domain knowledge in KG. Furthermore, it can be observed from the Medicine\_NER in Table \ref{table_domain} that the performance improvement using the MedicalKG is very obvious. From these results, we can conclude that KG, especially the domain KG, is very helpful for domain-specific tasks.

\subsection{Ablation studies}

%\begin{figure}[htb]
%\centering
%\includegraphics[width=1.0\columnwidth]{./figures/ablation_studies} 
%\caption{Ablation studies on Medicine\_NER with MedicalKG.}
%\label{ablation_studies}
%\end{figure}

\begin{figure}[htb]
\centering
\subfigure[Law\_Q\&A]{
\label{ablation_studies_a}
\includegraphics[width=0.9\columnwidth]{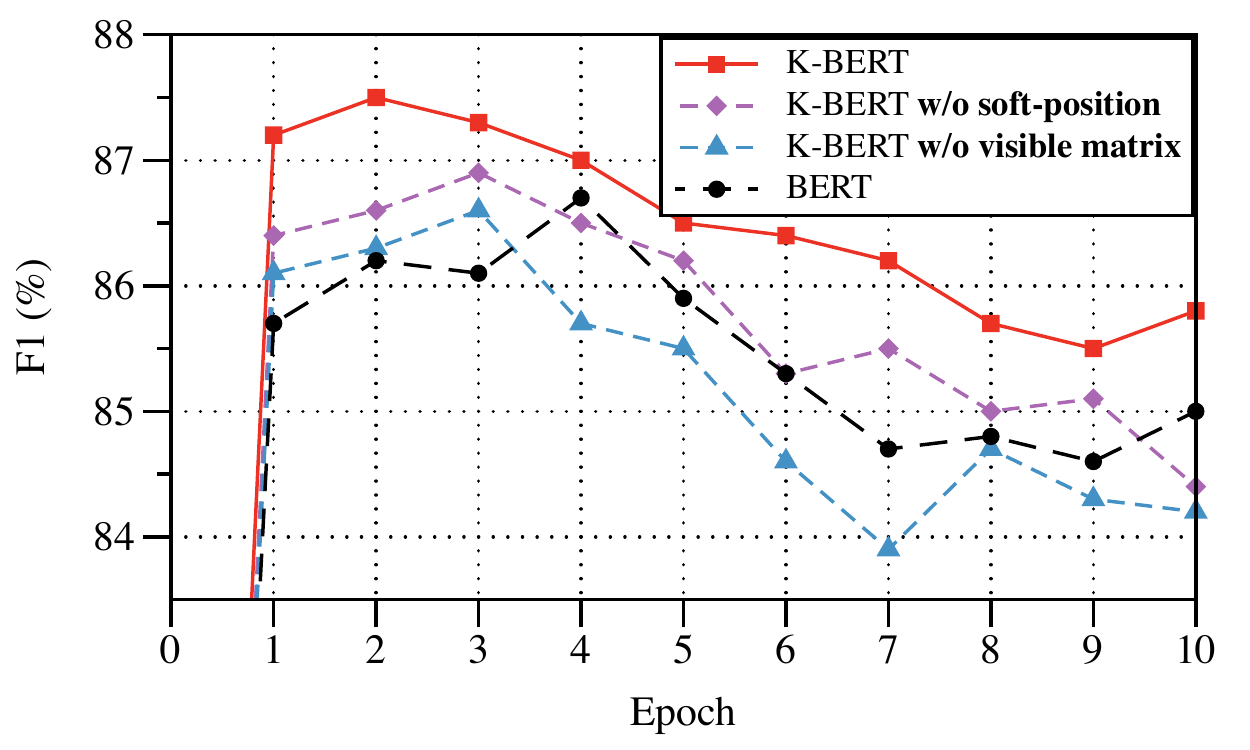}
}
\hspace{1in} 
\subfigure[Medicine\_NER]{
\includegraphics[width=0.9\columnwidth]{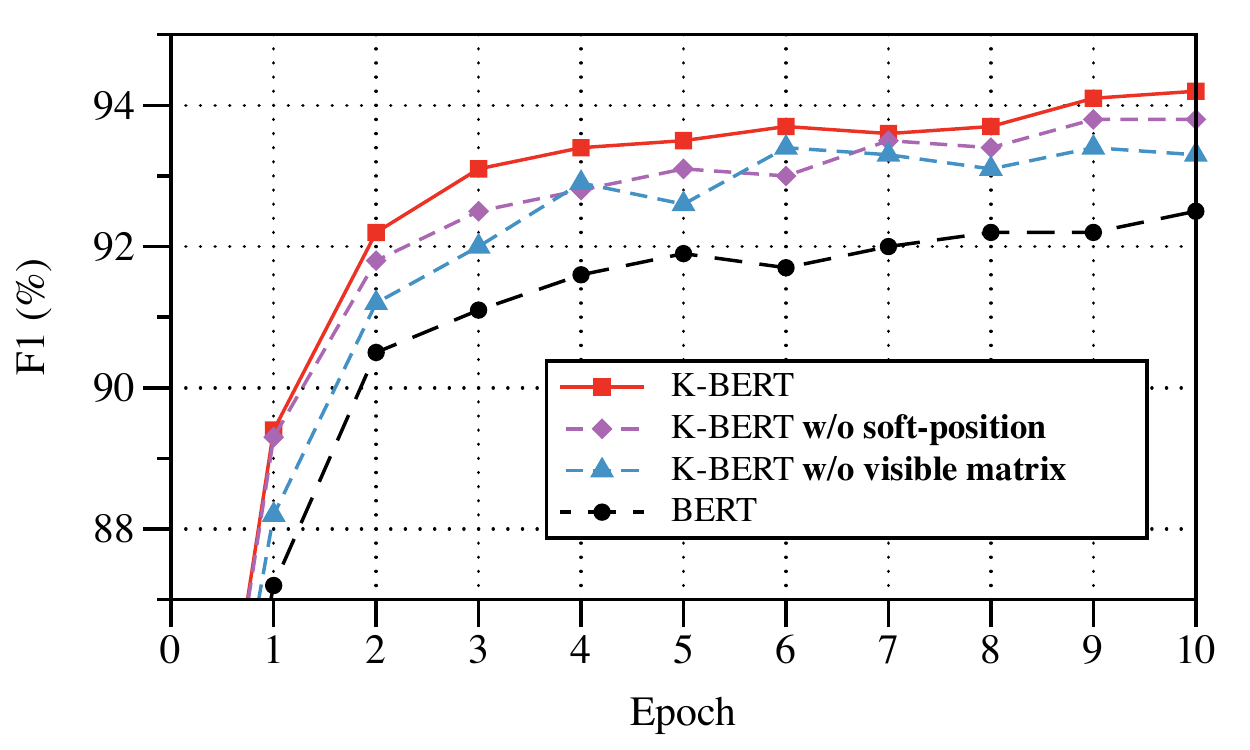}
}
\caption{Ablation studies: (a) Law\_Q\&A with CN-DBpedia; (b) Medicine\_NER with MedicalKG; }
\label{ablation_studies}
\end{figure}

In this subsection, we explore the effects of the soft-position and visible matrix for K-BERT using two domain-specific tasks (Law\_Q\&A and Medicine\_NER). ``\textbf{w/o soft-position}" refers to fine-tuning K-BERT with hard-position instead of soft-position. ``\textbf{w/o visible matrix}" means that the all tokens are visible to each other. BERT is equivalent to the K-BERT without KG. As shown in Figure \ref{ablation_studies}, we have the following observations: \textbf{(1)} In both tasks, without soft-position or visible matrix, the performance of K-BERT has declined; \textbf{(2)} In Law\_Q\&A (Figure \ref{ablation_studies_a}), K-BERT without visible matrix is worse than BERT, which confirms the existence of KN, i.e., improperly adding knowledge can lead to performance degradation; \textbf{(3)} In Figure \ref{ablation_studies_a}, K-BERT reaches its peak at epoch 2, while BERT is at epoch 4, which proves that K-BERT converges faster than BERT. In general, we can conclude that \textbf{the soft-position and the visible matrix can make K-BERT more robust to KN interference and thus make more efficient use of knowledge.}

\section{Conclusions}
\label{sec:conclusions}

% 在本文中，我们提出K-BERT使用知识图表进行语言表示，实现常识或领域知识的能力。具体来说，K-BERT首先将KG的知识注入句子，使其成为知识丰富的句子树。 接下来，软位置和可见矩阵适用于控制知识的范围，防止其意义变化。

In this paper, we propose K-BERT to enable language representation with knowledge graphs, achieving the capability of commonsense or domain knowledge. Specifically, K-BERT first injects knowledge from KG into a sentence, making it a knowledge-rich sentence tree. Next, soft-position and visible matrix are adapted to control the scope of knowledge, preventing it from deviating from its original meaning.

Despite the challenges of HES and KN, our investigation reveals promising results on twelve open-/specific- domain NLP tasks. Empirical results demonstrate that KG is especially helpful for knowledge-driven specific-domain tasks and can be used to solve problems that require domain experts. Besides, K-BERT is compatible with the model parameters of BERT, which means that users can directly adopt the available pre-trained BERT parameters (e.g., Google BERT, Baidu-ERNIE, etc.) on K-BERT without pre-training by themselves.

These positive results point to future work in \textbf{(1)} improving K-Query to enable filtering of unimportant triples based on context; \textbf{(2)} extending this approach to other LR models such as ELMo \cite{peters2018deep}, XLNet \cite{yang2019xlnet}, etc;

\section{Acknowledgement}

This work is funded by 2019 Tencent Rhino-Bird Elite Training Program.

%\newpage
\bibliographystyle{aaai} 
\bibliography{references.bib}

\end{document}